\documentclass{article}

\PassOptionsToPackage{numbers, compress}{natbib}

     \usepackage[final]{tackling_climate_workshop_style}

\usepackage[utf8]{inputenc} 
\usepackage[T1]{fontenc}    
\usepackage{hyperref}       
\usepackage{url}            
\usepackage{booktabs}       
\usepackage{amsfonts}       
\usepackage{nicefrac}       
\usepackage{microtype}      

\usepackage{amsmath, amssymb, amsthm}
\usepackage{graphicx}
\usepackage{algorithm}
\usepackage{algpseudocode}

\algtext*{EndFor}
\algtext*{EndWhile}

\newtheorem{theorem}{Theorem}[section]  

\title{Training-Free Data Assimilation with GenCast}

\author{%
  Thomas Savary \\
  ENS Paris-Saclay\\
  \texttt{thomas.savary@ens-paris-saclay.fr} \\
  \And
    François Rozet \\
    University of Liège \\
    \texttt{francois.rozet@uliege.be} \\
  \And
    Gilles Louppe \\
    University of Liège \\
    \texttt{g.louppe@uliege.be}
}

\begin{document}

\maketitle

\begin{abstract}
  Data assimilation is 
  widely used in many disciplines such as meteorology, oceanography, and robotics
  to estimate the state of a dynamical system from noisy observations. 
  In this work, we propose a lightweight and general method to perform data assimilation using diffusion models pre-trained for emulating dynamical systems. Our method builds on particle filters, a class of data assimilation algorithms, and does not require any further training. As a guiding example throughout this work, we illustrate our methodology on GenCast, a diffusion-based model that generates global ensemble weather forecasts.
\end{abstract}

\section{Introduction} \label{section:Introduction}
Simulating physical phenomena traditionally involves solving partial differential equations with dedicated numerical solvers \cite{ARAKAWA1977173, NavierStokesSimulation, BirdsallLangdon}. Recently, neural networks \cite{lecun2015deep} have emerged as a compelling alternative, achieving competitive accuracy at substantially lower computational cost \cite{pmlr-v70-tompson17a, sanchez2020learning, doi:10.1126/science.adi2336}. In particular, diffusion models \cite{pmlr-v37-sohl-dickstein15, DDPM} attract growing interest due to their ability to capture high-dimensional, multimodal distributions \cite{rombach2022highresolutionimagesynthesislatent}, making them promising candidates for modeling physical systems \cite{10.5555/3737916.3738648, NEURIPS2023_8df90a14, rozet2025lostlatentspaceempirical, price2025probabilistic}. In this context, the present paper contributes to ongoing efforts to adapt data assimilation algorithms to this new diffusion-based paradigm \cite{10.5555/3692070.3692867, sun2025aligndaalignscorebasedatmospheric, andry2025appabendingweatherdynamics, NonlinearEnsembleFilteringwithDiffusionModelsApplicationtotheSurfaceQuasigeostrophicDynamics}.

More precisely, we consider the filtering problem, which aims to estimate the state of a dynamical system at time $k$ from past and present observations $y^{1:k}$, that is, to approximate the posterior distribution $p(x^{k} \mid y^{1:k})$ \cite{DA}. For this purpose, we focus on particle filters \cite{PF}. Unlike commonly used data assimilation methods such as 4D-Var \cite{4D-VAR} or the Ensemble Kalman filter \cite{ENKF}, particle filters do not rely on linearizations that may not fully capture highly nonlinear processes, and thus represent a promising alternative. Since GenCast \cite{price2025probabilistic}, an autoregressive diffusion model for generating global weather predictions, perfectly matches our problem setting, we use it as a case study in what follows.

\section{Background} \label{section:Background}

\paragraph{GenCast and diffusion models}
GenCast \cite{price2025probabilistic} is a global, data-driven weather forecasting system based on diffusion models that produces 15-day ensemble forecasts at 12-hour intervals and 0.25° resolution. To achieve this, the globe’s surface is discretized into grid points, each described by a set of surface and atmospheric variables (e.g., temperature, humidity, wind components), and the model autoregressively generates probable future states from current states.

More precisly, given $x^{k}$ the approximation of the complete atmospheric state at time $k$ using surface and atmospheric variables at the different points of the grid, GenCast generates samples from the distribution $p(x^{k+1} \mid x^{k})$. To do so, as explained in \cite{price2025probabilistic,song2021scorebasedgenerativemodelingstochastic}, we solve a stochastic differential equation of the form 
\begin{equation} \label{eq:backward_sde_gencast}
    \mathrm{d}x^{k+1}_{t} = \left [ f_{t}x^{k+1}_{t} - \frac{1 + \eta^{2}}{2}g_{t}^{2} \nabla_{x^{k+1}_{t}} \log p\left(x^{k+1}_{t} \mid x^{k} \right) \right]\mathrm{d}t + \eta g_{t} \mathrm{d}w_{t}
\end{equation}
where $\eta \in \mathbb{R}_{+}$ is a parameter controlling stochasticity, $f_{t} \in \mathbb{R}$ is the drift coefficient, $g_{t} \in \mathbb{R}_{+}$ is the diffusion coefficient, $w_{t} \in \mathbb{R}^{n}$ denotes a standard Wiener process and $x^{k+1}_{t} \in \mathbb{R}^{n}$ is the sample perturbed with noise at time $t \in [0, 1]$ through a Gaussian kernel $p(x^{k+1}_{t} \mid x^{k+1}) = \mathcal{N}(\alpha_{t}x^{k+1}, \sigma^{2}_{t}I)$. The coefficients $\alpha_{t}$ and $\sigma_{t}$ are derived from $f_{t}$ and $g_{t}$ such that $p(x^{k+1}_{1} \mid x^{k+1}) \approx \mathcal{N}(0, \sigma_{1}^{2}I)$ \cite{song2021scorebasedgenerativemodelingstochastic}. 

The score function $\nabla_{x^{k+1}_{t}} \log p\left(x^{k+1}_{t} \mid x^{k} \right)$ in Equation \eqref{eq:backward_sde_gencast} is unknown in practice, but can be approximated by a neural network $d_{\theta}$ called denoiser and trained to minimize the denoising error \cite{EDM}. The optimal denoiser $d_{\theta^{*}}$ is the mean $\mathbb{E}\left [{x}^{k+1} \mid x^{k}, x^{k+1}_{t} \right]$ of $p({x}^{k+1} \mid x^{k}, x^{k+1}_{t})$ and is linked to the score function through Tweedie's formula \cite{Tweedie_1947, 04f9ad2d-909a-34da-ad09-b5ac91b665b6} (see Appendix \ref{appendix:Tweedie})
\begin{equation} \label{eq:Tweedie}
    \mathbb{E}\left [{x}^{k+1} \mid x^{k}, x^{k+1}_{t} \right] = \alpha_{t}^{-1} \left[ x^{k+1}_{t} + \sigma_{t}^{2} \nabla_{x^{k}_{t}} \log p(x_{t}^{k+1} \mid x^{k}) \right]
\end{equation}
which allows to use $s_{\theta}(x^{k+1}_{t}, x^{k}, t) = \sigma_{t}^{-2} \left[\alpha_{t}d_{\theta} \left(x^{k+1}_{t}, x^{k}, t\right) - x^{k+1}_{t}\right]$ as a score estimate in Equation \eqref{eq:backward_sde_gencast}. Thus, drawing noise samples from $p(x^{k+1}_{1} \mid x^{k}) \approx \mathcal{N}(0, \sigma_{1}^{2}I)$ and solving Equation \eqref{eq:backward_sde_gencast} from $t= 1$ to $0$ with an appropriate discretization scheme \cite{EDM, lu2023dpmsolverfastsolverguided}, we obtain samples from $p(x^{k+1} \mid x^{k})$.

\paragraph{Particle filters}
The aim of particle filters \cite{PF} is to approximate the filtering posterior distribution $p(x^{k} \mid y^{1:k})$ by a finite discrete probability measure $ \mu^{k}_{x} = \sum_{i = 1}^{N} w^{k}_{i}\delta_{x^{k}_{i}}$ such that the following converges weakly
\begin{equation} \label{eq:weak_convergence}
    \sum_{i = 1}^{N} w^{k}_{i}g(x^{k}_{i}) \underset{N \to +\infty}{\longrightarrow} \int g(x^{k})p(x^{k} \mid y^{1:k}) \mathrm{d}x^{k} 
\end{equation}
where $x^{k}_{i}$ are the particles at time step $k$, $w^{k}_{i}$ the associated weights, $y^{1:k}$ the observations and $g$ a continuous and bounded function. In their standard form, particle filters alternate between a sampling step, where particles are propagated using a proposal distribution, and a weighting step, where weights are updated according to the proposal. 

In this work, we focus on the "optimal" proposal $p(x^{k+1} \mid x^{k}, y^{k+1})$ \cite{PerformanceBoundsforParticleFiltersUsingtheOptimalProposal}, which propagates particles from step $k$ to $k+1$ conditionally on the next observation $y^{k+1}$. This proposal is coined "optimal" as it minimizes the variance of the weights, which can then be computed recursively as
\begin{equation} \label{eq:weights_optimal}
    w^{k+1}_{i} := p(y^{k+1} \mid x^{k}_{i}) \times w^{k}_{i}. 
\end{equation}
The main drawback of particle filters is the degeneracy of the algorithm, which corresponds to the situation where only a small subset of particles have non-negligible weights. This is due to the dimension of the observation space: the higher this dimension, the more peaked
the likelihood is, and the more unlikely it is for the majority of particles to end up close to all the observations \cite{reading50238}.

\section{Methodology} \label{section:Methodology}

\paragraph{Sampling from the optimal proposal distribution}
The use of the optimal proposal distribution suggests that we are able to draw samples from $p(x^{k+1} \mid x^{k}, y^{k+1})$, which is not often the case in practice. However, for diffusion models like GenCast, this can be
done relatively easily by using the posterior score $\nabla_{x^{k+1}_{t}} \log p\left(x^{k+1}_{t} \mid x^{k}, y^{k+1} \right)$ when solving Equation \eqref{eq:backward_sde_gencast} \cite{andry2025appabendingweatherdynamics,song2021scorebasedgenerativemodelingstochastic}.

Thanks to Bayes’ rule, the posterior score $\nabla_{x^{k+1}_{t}} \log p\left(x^{k+1}_{t} \mid x^{k}, y^{k+1} \right)$ can be decomposed into two terms as \cite{song2021scorebasedgenerativemodelingstochastic, chung2024diffusionposteriorsamplinggeneral}
\begin{equation} \label{eq:posterior_score}
    \nabla_{x^{k+1}_{t}} \log p\left(x^{k+1}_{t} \mid x^{k}, y^{k+1} \right) = \nabla_{x^{k+1}_{t}} \log p\left(x^{k+1}_{t} \mid x^{k} \right) + \nabla_{x^{k+1}_{t}} \log p\left(y^{k+1} \mid x^{k+1}_{t}, x^{k} \right).
\end{equation}
As an estimate of the first term is already available via the pre-trained denoiser (see Section \ref{section:Background}), the remaining task is to estimate the likelihood score $\nabla_{x^{k+1}_{t}} \log p\left(y^{k+1} \mid x^{k+1}_{t}, x^{k} \right)$. To do so, assuming a differentiable observation operator $\mathcal{H}$, a diagonal covariance matrix $\Sigma_{y}$  for the observations and a Gaussian forward process $p(y^{k+1} \mid x^{k+1}) = \mathcal{N}(y^{k+1} \mid \mathcal{H}(x^{k+1}), \Sigma_{y})$, we evaluate the likehood score as \cite{MMPS}
\begin{equation} \label{eq:MMPS_estimation}
    \nabla_{x^{k+1}_{t}} \log p(y^{k+1} \mid x^{k+1}_{t}, x^{k}) = \nabla_{x^{k+1}_{t}} \mathbb{E}[x^{k+1} \mid x^{k+1}_{t}, x^{k}]^{T}H^{T} \left(\Sigma_{y} + H V H^{T} \right)^{-1} v^{k+1}
\end{equation}
where $H$ is the Jacobian of $\mathcal{H}$, $V = \mathbb{V}[x^{k+1} \mid x^{k+1}_{t}, x^{k}]$ and $v^{k+1} = y^{k+1} - \mathcal{H}(\mathbb{E}[x^{k+1} \mid x^{k+1}_{t}, x^{k}])$. Despite its complex form, this term can be computed efficiently via automatic differentiation and using a linear solver (see \cite{MMPS} or Appendix \ref{appendix:MMPS} for more details). 

\paragraph{Computing weights}
Updating the weights in the case of the optimal proposal is non-trivial because we cannot evaluate $p(y^{k+1} \mid x^{k})$ directly in Equation \eqref{eq:weights_optimal}. We then propose to approximate $p(x^{k+1} \mid x^{k})$ by a Dirac distribution \cite{GVK164761632} at $\mathbb{E}[x^{k+1} \mid x^{k}]$ so that
\begin{equation} \label{eq:weights_approximation}
    p(y^{k+1} \mid x^{k}) = \int p(y^{k+1} \mid x^{k+1}) p(x^{k+1} \mid x^{k}) \mathrm{d}x^{k+1} \approx p(y^{k+1} \mid \mathbb{E}[x^{k+1} \mid x^{k}]).
\end{equation}
The conditional expectation $\mathbb{E}[x^{k+1} \mid x^{k}]$ is not known a priori, but can be efficiently estimated using the pre-trained denoiser
\begin{equation} \label{eq:approx_E_denoiser}
    \mathbb{E}[x^{k+1} \mid x^{k}]  \underset{\varepsilon \sim \mathcal{N}(0,I)}{=}  \mathbb{E}[x^{k+1} \mid x^{k}, \sigma_{1}\varepsilon] \approx d_{\theta}\left(x^{k+1}_{t=1} = \sigma_{1}\varepsilon, x^{k}, t = 1 \right)
\end{equation}

These two elements enable the use of the Fully-Adapted Auxiliary Particle Filter (FA-APF), a particle filter algorithm adapted to the optimal proposal \cite{FA-APF} and described in Algorithm \ref{algo:FA-APF}. We introduce an inflation coefficient $\alpha$ to control the degeneracy of the weights, at the expense of a bias in the approximation of the posterior distribution $p(x^{k} \mid y^{1:k})$.

\begin{algorithm}
    \caption{Fully-Adapted Auxiliary Particle Filter (FA-APF)}
    \begin{algorithmic}[1]

        \Require initial condition $x^{0}$, number of particles $N$, thresholds $N_{\text{thr}}^{\text{min}, \text{max}}$, number of steps $K$.
        \State $x^{0}_{i} \gets x^{0}$
        \State $w^{0}_{i} \gets 1/N$
        \For{$k$ in $0,...,K-1$}
            \State $\mu^{k+1}_{i} \gets \mathbb{E}[x^{k+1} \mid x^{k}_{i}]$
            \While{$N_{\text{eff}}$ not in $[N_{\text{thr}}^{\text{min}}, N_{\text{thr}}^{\text{max}}]$}
                \State update/initialize $\alpha$
                \State $\hat{w}^{k+1}_{i} \gets \left[ p(y^{k+1} \mid \mu^{k+1}_{i})\right]^{\alpha}$
                \State $w^{k+1}_{i} \gets \hat{w}^{k+1}_{i} / \sum_{j=1}^{N} \hat{w}^{k+1}_{j}$
                \State $N_{\text{eff}} \gets 1 / \sum_{i=1}^{N}(w^{k+1}_{i})^{2}$
            \EndWhile
            \State $a^{k+1}_{i} \sim \text{Cat}(\{w^{k+1}_{i}\}_{1 \leq i \leq N})$
            \State $x^{k+1}_{i} \gets p(x^{k+1} \mid x^{k}_{a^{k+1}_{i}}, y^{k+1})$
        \EndFor
        \State \Return $\mu^{k}_{x} = \frac{1}{N}\sum_{i = 1}^{N}\delta_{x^{k}_{i}}$ for all $k \in [1,K]$
    \end{algorithmic}
    \label{algo:FA-APF}
\end{algorithm}

\section{Results}
\paragraph{Experimental setup}To evaluate our method, we apply Algorithm \ref{algo:FA-APF} with the pre-trained GenCast denoiser at 1° resolution and $N=256$ particles ($N_{\mathrm{thr}}^{\mathrm{min}}=60$, $N_{\mathrm{thr}}^{\mathrm{max}}=70$). The observations $y^{k}$ and the initial condition $x^{0}$ are obtained from a reference ERA5 trajectory (a global atmospheric reanalysis produced by ECMWF \cite{ERA5}). We only observe temperature at the surface and at all pressure levels on a regular latitude–longitude grid, taking one point every four degrees in both directions, with zero-mean Gaussian noise and a standard deviation of $0.1$ Kelvin. This setup corresponds to a linear observation operator $\mathcal{H}$ and a covariance matrix $\Sigma_{y} = (0.1)^{2}I$ for the observations. Following \cite{rozet2025lostlatentspaceempirical}, we solve Equation \eqref{eq:backward_sde_gencast} using a temporal discretization of 40 time steps and a third-order Adam-Bashforth scheme \cite{Hairer} with two correction steps  \cite{song2021scorebasedgenerativemodelingstochastic} and two BiCGStab \cite{BiCGStab} iterations to solve the linear system of Equation \eqref{eq:MMPS_estimation}.

\begin{figure}[!h]
  \centering
  \includegraphics[width=.6\textwidth]{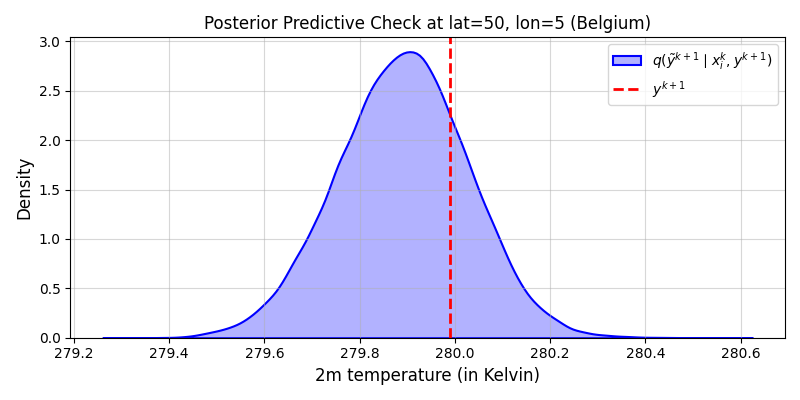}
  \caption{Conditional posterior predictive distribution (blue curve) and corresponding observation (red dashed line) at an arbitrarily chosen point of the grid with coordinates (lat=50, lon=5) for the surface temperature variable. The observation is consistent with the posterior predictive distribution.}
  \label{fig:ppc}
\end{figure}

First, to validate the correctness of the sampling from the optimal proposal (see Section \ref{section:Methodology}), we verify the consistency of the observations $y^{k+1}$ with the conditional posterior predictive distribution $q(\tilde{y}^{k+1} \mid x^{k}_{i}, y^{k+1}) = \mathbb{E}_{q(x^{k+1} \mid x^{k}_{i}, y^{k+1})} \left[ p(\tilde{y}^{k+1} \mid x^{k+1}) \right]$. An example for a specific variable (surface temperature) at an arbitrarily chosen point of the grid is shown in Figure \ref{fig:ppc}.

\begin{figure}[!h]
  \centering
  \includegraphics[width=1.\textwidth]{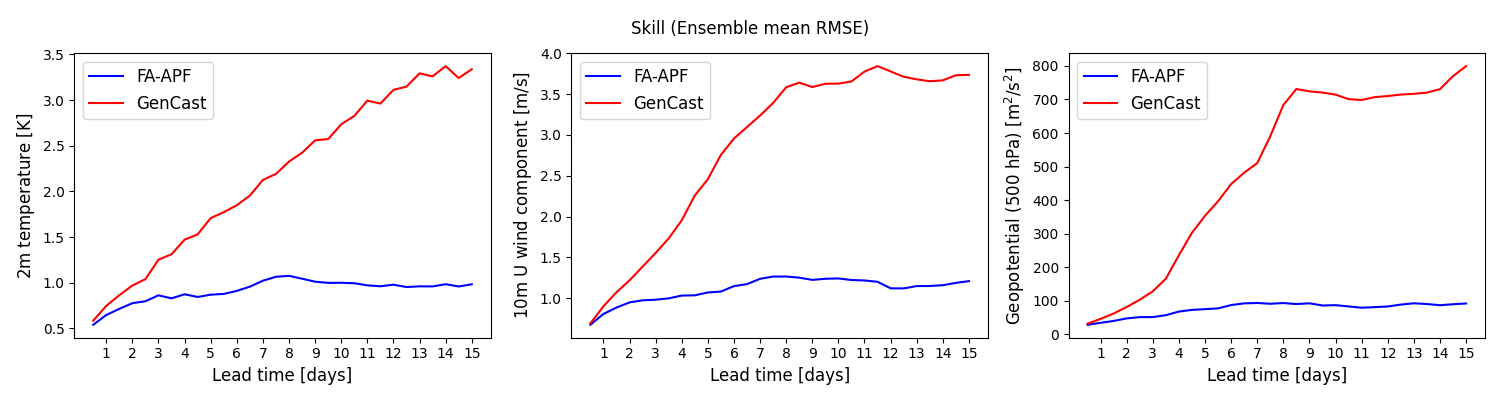}
  \caption{Skill comparison between the FA-APF (blue curve) and the ensemble of unconditional GenCast trajectories (red curve) for the surface temperature (left), the surface U component of wind (middle) and the geopotential at 500 hPA (right). The FA-APF allows to obtain a low and more or less constant skill after 7 days of observations, even for unobserved variables.}
  \label{fig:skill}
\end{figure}

Then, we compare the skill (RMSE of the ensemble mean, lower the better) at each time step for the FA-APF and an ensemble of $N=256$ unconditional forecasts generated autoregressively using Equation \eqref{eq:backward_sde_gencast} without conditioning the score on observations. The skill is computed using the reference ERA5 trajectory from which observations are extracted as ground truth. Figure \ref{fig:skill} shows that the skill of the filter reaches a plateau for all variables (including unobserved ones), which is well below the skill of unconditional trajectories. Further results, including skill scores for additional variables, ensemble spread, and trajectory visualizations are presented in Appendix \ref{appendix:sup_results}.

\section{Conclusion}
This work introduces a training-free data assimilation method that shows promising results with GenCast and could be deployed in operational settings with minimal effort. Since it requires no additional training, the approach is readily applicable to other autoregressive diffusion models, and extending its evaluation beyond GenCast is a natural next step \cite{rozet2025lostlatentspaceempirical, larsson2025diffusionlam, sea-ice}. 

Future work will investigate the role of the initial condition $x^{0}$, the thresholds $N_{\text{thr}}^{\text{min,max}}$, and the number of particles $N$ used as hyperparameters in Algorithm \ref{algo:FA-APF}. Another research direction is to extend this work toward a training-free approach to the more complex problem of reanalysis, which seeks to estimate the state of a dynamical system from past, present, and future observations \cite{DA}.

\section*{Acknowledgments and Disclosure of Funding}
François Rozet is a research fellow of the F.R.S.-FNRS (Belgium) and acknowledges its financial support.

We gratefully acknowledge the support of NVIDIA for supporting this research project through the NVIDIA Academic Grant Program.

\bibliographystyle{unsrtnat}
\bibliography{reference}

\newpage
\appendix
\section{Tweedie’s formulae} \label{appendix:Tweedie}

\begin{theorem}
Assuming that $p(x^{k}_{t}\mid x^{k}) = \mathcal{N}(x^{k}_{t} \mid \alpha_{t} x^{k}, \sigma_{t}^{2}I)$ and that $x^{k+1}_{t}$ is conditionally independent of $x^{k}$ given $x^{k+1}$, the first moment of the distribution $p(x^{k+1}_{t} \mid x^{k})$ is linked to the score function $\nabla_{x^{k+1}_{t}} \log p(x^{k+1}_{t} \mid x^{k}) $ used in Equation \eqref{eq:backward_sde_gencast} through
\begin{equation}
    \nabla_{x^{k+1}_{t}} \log p(x^{k+1}_{t} \mid x^{k}) = \sigma_{t}^{-2}\left( \alpha_{t} \mathbb{E}[x^{k+1} \mid x^{k+1}_{t}, x^{k}]  - x^{k+1}_{t} \right)   
\end{equation}
\end{theorem}
We provide proofs of this theorem for completeness, even though it is a well known result \cite{Tweedie_1947, 04f9ad2d-909a-34da-ad09-b5ac91b665b6}.
\begin{proof}
    \setlength{\jot}{10pt}
    \begin{align*}
        \nabla_{x^{k+1}_{t}} \log p(x^{k+1}_{t} \mid x^{k}) &= \frac{1}{p(x^{k+1}_{t} \mid x^{k})} \nabla_{x^{k+1}_{t}} p(x^{k+1}_{t} \mid x^{k}) \\
        &= \frac{1}{p(x^{k+1}_{t} \mid x^{k})} \int \nabla_{x^{k+1}_{t}} p(x^{k+1}_{t}, x^{k+1}\mid x^{k}) \mathrm{d}x^{k+1} \\
        &= \frac{1}{p(x^{k+1}_{t} \mid x^{k})} \int p(x^{k+1}_{t}, x^{k+1}\mid x^{k}) \nabla_{x^{k+1}_{t}} \log p(x^{k+1}_{t}, x^{k+1}\mid x^{k}) \mathrm{d}x^{k+1} \\
        &= \int p(x^{k+1} \mid x^{k+1}_{t}, x^{k}) \nabla_{x^{k+1}_{t}} \log p(x^{k+1}_{t} \mid x^{k+1}) \text{d}x^{k+1} \\
        &= \int p(x^{k+1} \mid x^{k+1}_{t}, x^{k}) \sigma_{t}^{-2} \left( \alpha_{t} x^{k+1} - x^{k+1}_{t} \right) \mathrm{d}x^{k+1} \\
        &= \alpha_{t} \sigma_{t}^{-2} \int x^{k+1} p(x^{k+1} \mid x^{k+1}_{t}, x^{k}) \mathrm{d}x^{k+1} - \sigma_{t}^{-2} x^{k+1}_{t} \int p(x^{k+1} \mid x^{k+1}_{t}, x^{k}) \mathrm{d}x^{k+1} \\
        &= \alpha_{t}\sigma_{t}^{-2} \mathbb{E}[x^{k+1} \mid x^{k+1}_{t}, x^{k}] - \sigma_{t}^{-2} x^{k+1}_{t} \\
        &= \sigma_{t}^{-2}\left( \alpha_{t} \mathbb{E}[x^{k+1} \mid x^{k+1}_{t}, x^{k}]  - x^{k+1}_{t} \right)
    \end{align*}
\end{proof}

\section{Moment Matching Posterior Sampling} \label{appendix:MMPS}

We provide technical details on how $\nabla_{x^{k+1}_{t}} \log p(y^{k+1} \mid x^{k+1}_{t}, x^{k})$ is estimated for completeness, even though it is already explained in \cite{MMPS}.

In order to generate samples conditionally on $y^{k+1}$, we need to evaluate $\nabla_{x^{k+1}_{t}} \log p(y^{k+1} \mid x^{k+1}_{t}, x^{k})$ and plug it into equation \eqref{eq:posterior_score}. To do so, we can first write  $p(y^{k+1} \mid x^{k+1}_{t}, x^{k})$ as an integral 
\begin{align} \label{eq:int_MM}
    p(y^{k+1} \mid x^{k+1}_{t}, x^{k}) &= \int p(y^{k+1}, x^{k+1} \mid x^{k+1}_{t}, x^{k}) \mathrm{d}x^{k+1} \\
    &= \int p(y^{k+1} \mid x^{k+1}) p(x^{k+1} \mid x^{k+1}_{t}, x^{k}) \mathrm{d}x^{k+1}
\end{align}
Then, assuming a differentiable observation operator $\mathcal{H}$, a diagonal covariance matrix $\Sigma_{y}$ for the observations, a Gaussian forward process $p(y^{k+1} \mid x^{k+1}) = \mathcal{N}( y^{k+1} \mid \mathcal{H}(x^{k+1}), \Sigma_{y})$ and a Gaussian approximation $q(x^{k+1} \mid x^{k+1}_{t}, x^{k}) = \mathcal{N}(x^{k+1} \mid \mathbb{E}[x^{k+1} \mid x^{k+1}_{t}, x^{k}], \mathbb{V}[x^{k+1} \mid x^{k+1}_{t}, x^{k}])$ of $p(x^{k+1} \mid x^{k+1}_{t}, x^{k})$, we obtain the following approximation 
\begin{align} \label{eq:approx_MM}
    q(y^{k+1} \mid x^{k+1}_{t}, x^{k})
    &= \int p(y^{k+1} \mid x^{k+1}) q(x^{k+1} \mid x^{k+1}_{t}, x^{k}) \mathrm{d}x^{k+1} \\
    &= \mathcal{N}\left( y^{k+1} \mid \mathcal{H}(\mathbb{E}[x^{k+1} \mid x^{k+1}_{t}, x^{k}]), \Sigma_{y} + H \mathbb{V}[x^{k+1} \mid x^{k+1}_{t}, x^{k}] H^{T} \right)
\end{align}
where $H$ is the Jacobian of $\mathcal{H}$. This approximation allows to estimate $\nabla_{x^{k+1}_{t}} \log p(y^{k+1} \mid x^{k+1}_{t},x^{k})$, under the assumption that the derivative of $\mathbb{V}[x^{k+1} \mid x^{k+1}_{t}, x^{k}]$ with respect to $x^{k+1}_{t}$ is negligible, as 
\begin{equation} \label{eq:approx_grad_MM_appendix}
    \nabla_{x^{k+1}_{t}} \log q(y^{k+1} \mid x^{k+1}_{t}, x^{k}) = \nabla_{x^{k+1}_{t}} \mathbb{E}[x^{k+1} \mid x^{k+1}_{t}, x^{k}]^{T}H^{T} \left(\Sigma_{y} + H V H^{T} \right)^{-1} v^{k+1}
\end{equation}
where $v^{k+1} = y^{k+1} - \mathcal{H}(\mathbb{E}[x^{k+1} \mid x^{k+1}_{t}, x^{k}])$ and $V = \mathbb{V}[x^{k+1} \mid x^{k+1}_{t}, x^{k}]$. Although Equation \eqref{eq:approx_grad_MM_appendix} gives an explicit formula to estimate $\nabla_{x^{k+1}_{t}} \log p(y^{k+1} \mid x^{k+1}_{t},x^{k})$, solving it in practice is not trivial. Indeed, if the dimension of system state is large, compute and store $\mathbb{V}[x^{k+1} \mid x^{k+1}_{t}, x^{k}]$ is impossible. However, as $\Sigma_{y} + H \mathbb{V}[x^{k+1} \mid x^{k+1}_{t}, x^{k}] H^{T}$ is symmetric
positive definite (SPD), we can apply the conjugate gradient method. This method is an iterative algorithm to solve linear systems of form $Mv = b$ (where $M$ is SPD), using only implicit access to $M$ through a matrix-vector operator. In our case, the linear system to solve is
\begin{align} \label{eq:CG_method_MM}
    v^{k+1} &= \left( \Sigma_{y} + H \mathbb{V}[x^{k+1} \mid x^{k+1}_{t}, x^{k}] H^{T} \right) v \\
    &= \Sigma_{y}v + \alpha_{t}^{-1} \sigma_{t}^{2}H(\underbrace{v^{T}H\nabla_{x^{k+1}_{t}} \mathbb{E}[x^{k+1} \mid x^{k+1}_{t}, x^{k}]}_{\text{vector-Jacobian product}})^{T}
\end{align}
Within automatic differentiation frameworks, the vector-Jacobian product on the right-hand side can be cheaply evaluate using the pre-trained denoiser as an estimator of $\mathbb{E}[x^{k+1} \mid x^{k+1}_{t}, x^{k}]$.

\section{Supplementary results} \label{appendix:sup_results}

\subsection{Skill}

\begin{figure}[!h]
  \centering
  \includegraphics[width=0.95\textwidth]{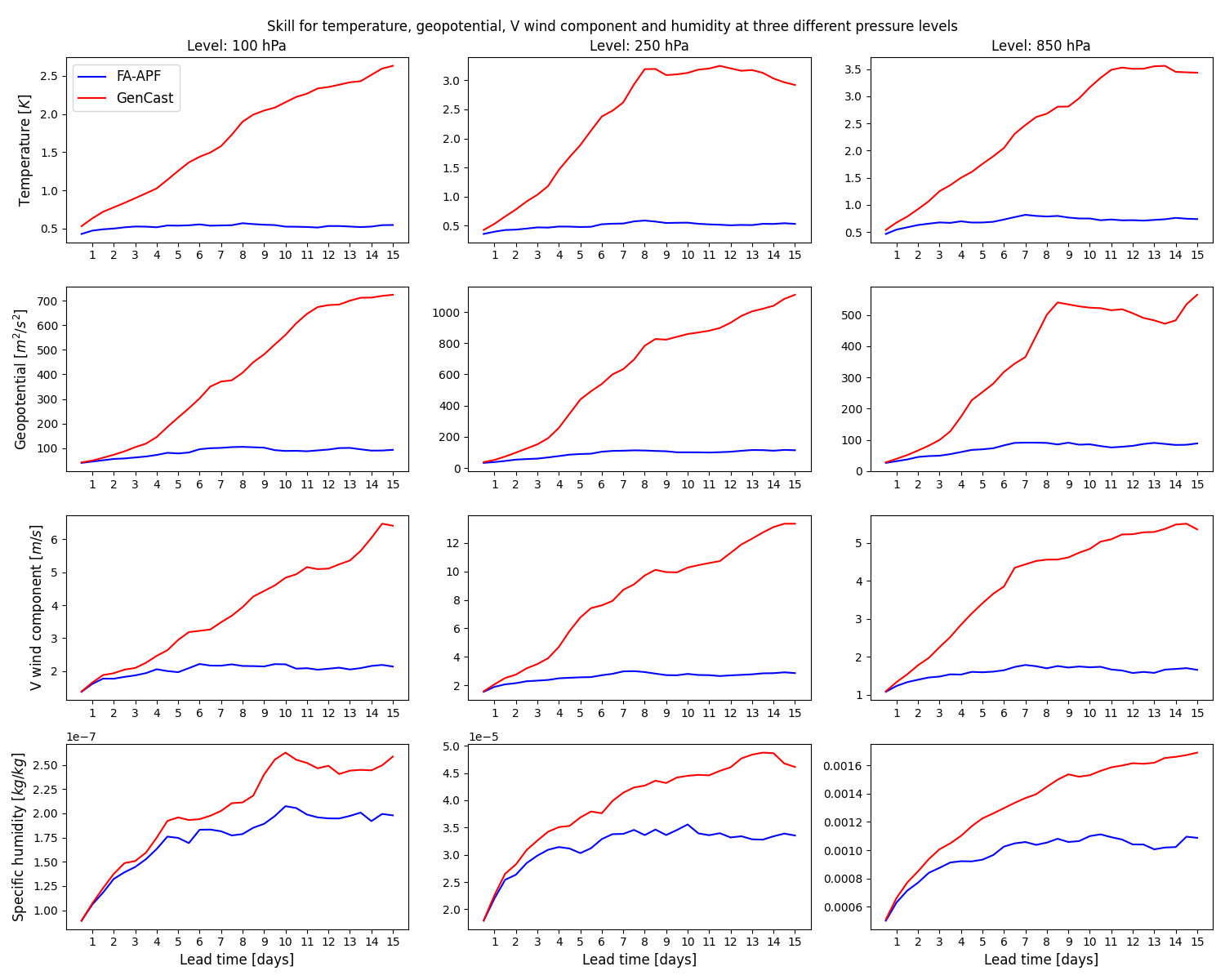}
  \caption{Skill for temperature, geopotential, V component of wind and specific humidity at three different pressure levels (100, 250 and 850 hPa). The skill reaches a plateau after a certain number of time steps for all variables (even those that are not observed), well below the one of GenCast's forecasts.}
  \label{fig:skill_appendix}
\end{figure}

\subsection{Spread}

\begin{figure}[!h]
  \centering
  \includegraphics[width=0.95\textwidth]{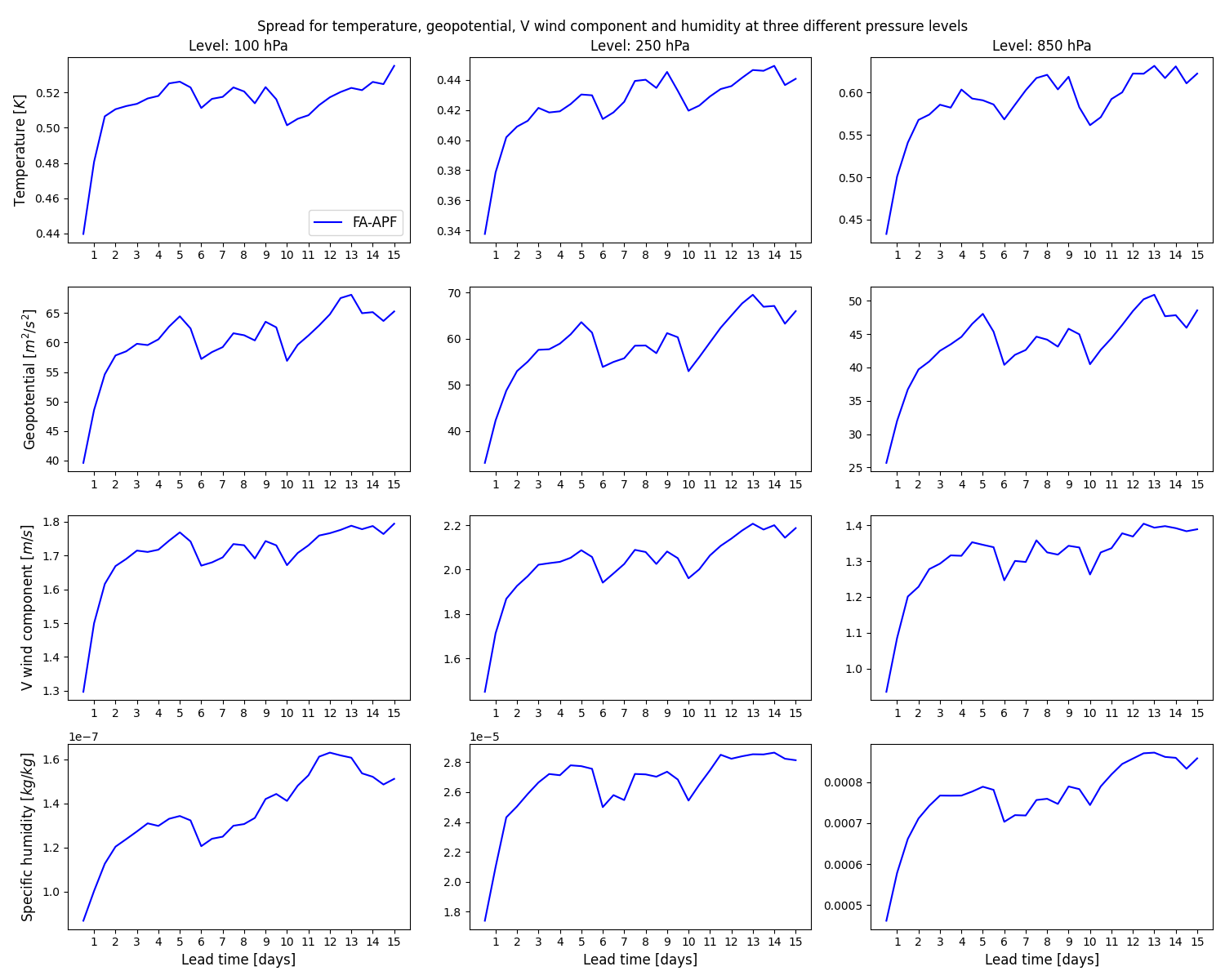}
  \caption{Spread for temperature, geopotential, V component of wind and specific humidity at three different pressure levels (100, 250 and 850 hPa). The spread is non-zero and of the same order of magnitude as the skill, indicating that we capture a distribution rather than collapsing onto a single mode.}
  \label{fig:spread_appendix}
\end{figure}

\subsection{Visualization of trajectories}

\begin{figure}[!h]
  \centering
  \includegraphics[width=0.95\textwidth]{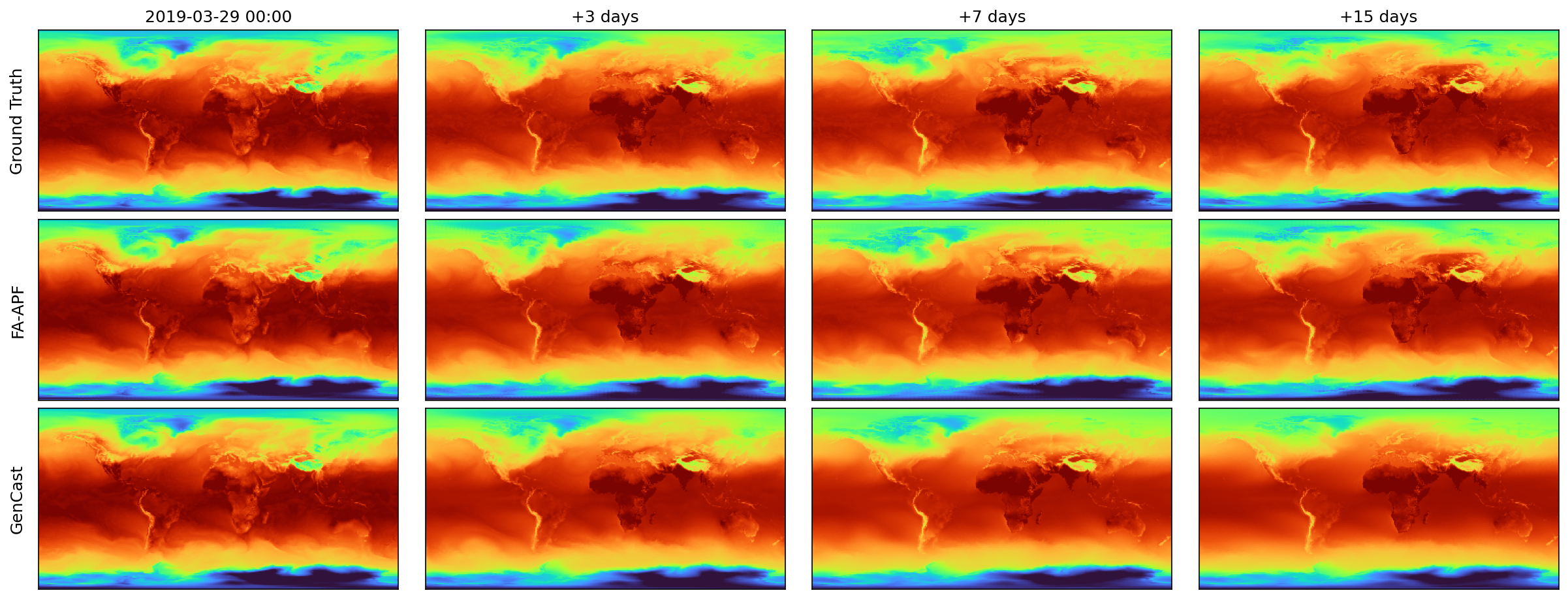}
  \caption{Comparison of surface temperature between the reference ERA5 trajectory (first row), the FA-APF ensemble mean (second row), and the GenCast ensemble mean (third row) after 3, 7, and 15 days.}
  \label{fig:traj_t2m}
\end{figure}

\begin{figure}[!h]
  \centering
  \includegraphics[width=0.95\textwidth]{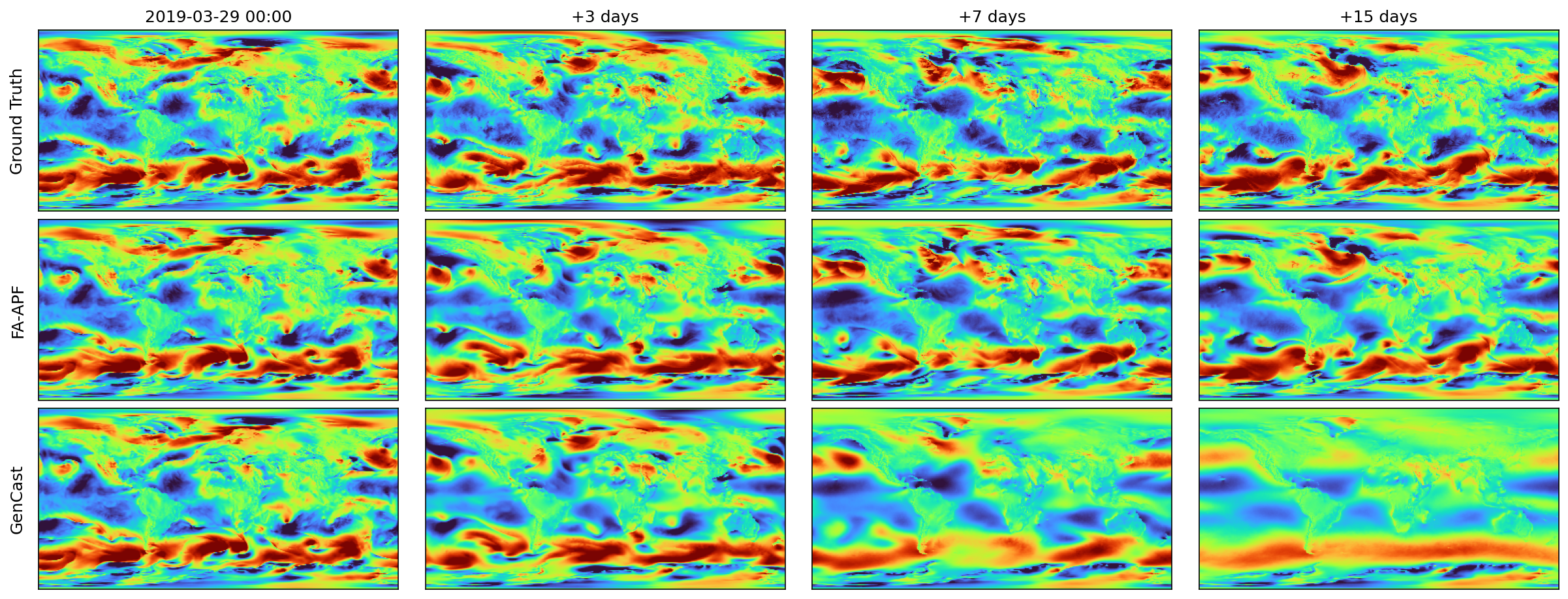}
  \caption{Comparison of the 10m U component of wind between the reference ERA5 trajectory (first row), the FA-APF ensemble mean (second row), and the GenCast ensemble mean (third row) after 3, 7, and 15 days.}
  \label{fig:traj_10m_u_wind}
\end{figure}

\begin{figure}[!h]
  \centering
  \includegraphics[width=0.95\textwidth]{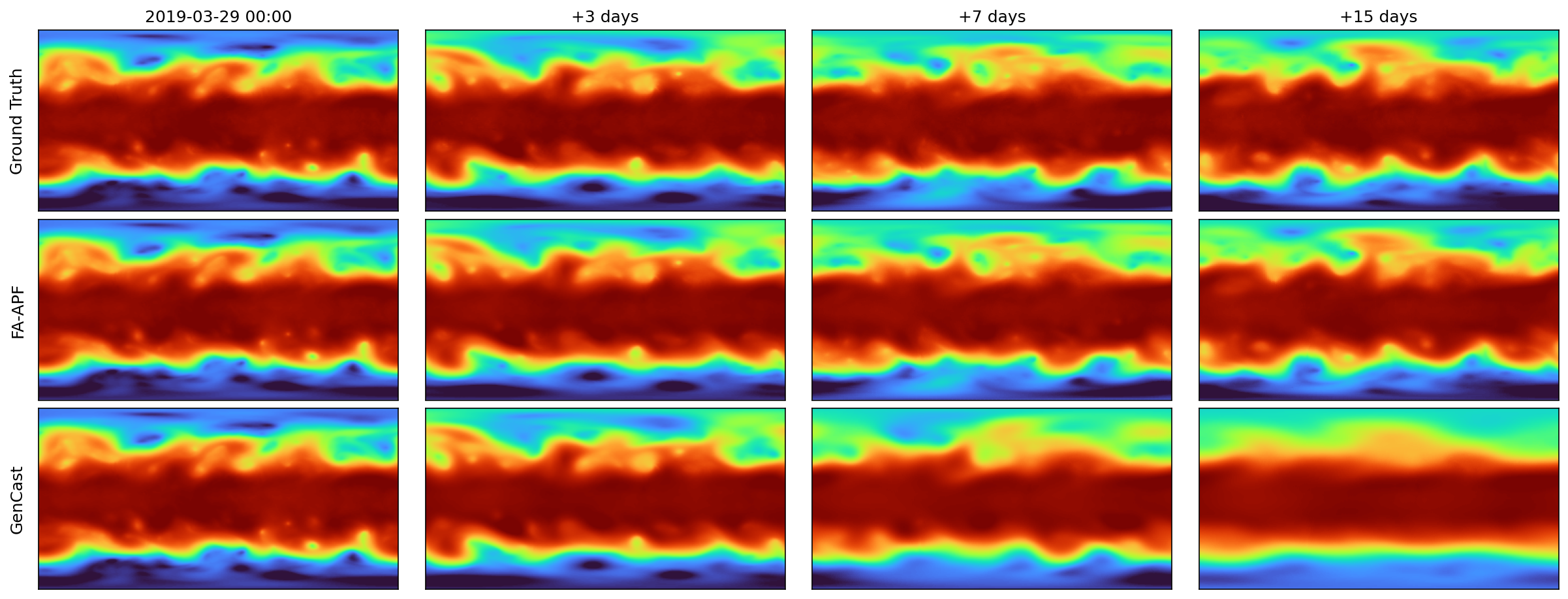}
  \caption{Comparison of the geopotential at 500 hPA between the reference ERA5 trajectory (first row), the FA-APF ensemble mean (second row), and the GenCast ensemble mean (third row) after 3, 7, and 15 days.}
  \label{fig:traj_z500}
  
\end{figure}

\end{document}